\documentclass[NewProceedings,letterpaper]{ascelike-new}
\usepackage[utf8]{inputenc}
\usepackage[T1]{fontenc}
\usepackage{lmodern}
\usepackage{graphicx}
\usepackage[style=base,figurename=Fig.,labelfont=bf,labelsep=period]{caption}
\usepackage{subcaption}
\usepackage{amsmath}
\usepackage{newtxtext,newtxmath}
\usepackage[colorlinks=true,citecolor=black,linkcolor=black]{hyperref}
\usepackage{makecell}
\usepackage{comment}

\makeatletter

\newcommand{\asce@authorline}{%
  Johannes Mootz\textsuperscript{1} and Reza Akhavian, Ph.D.\textsuperscript{2}%
}

\newcommand{\asce@affils}{%
\textsuperscript{1}Department of Civil, Construction, and Environmental Engineering, San Diego State University, San Diego, CA, United States; and Department of Mechanical and Aerospace Engineering, UC San Diego, San Diego, CA, United States. Email: jmootz@sdsu.edu\par
\textsuperscript{2}Associate Professor, Department of Civil, Construction, and Environmental Engineering, San Diego State University, San Diego, CA, United States (corresponding author). Email: rakhavian@sdsu.edu\par
}

\makeatletter

\renewcommand{\@maketitle}{%
  \newpage\null
  \ifthenelse{\boolean{Journal}}
    {\vspace{0.00in}}
    {\ifthenelse{\boolean{NewProceedings}}
      {\vspace{28pt}}
      {\vspace{1.40in}}}%
  \centering{\large\bfseries\@title\par}%
  \vspace{1.0em}%
  \centering{\normalsize\normalfont \asce@authorline\par}%
  \vspace{0.8em}%
  {\noindent\begin{minipage}{\textwidth}\raggedright\normalsize\normalfont
    \asce@affils
  \end{minipage}\par}%
  \vspace{0.5em}%
}

\makeatother

\begin{document}

\title{Control Barrier Functions with Audio Risk Awareness for Robot Safe Navigation on Construction Sites}

\maketitle

\begin{abstract}
Construction automation increasingly requires autonomous mobile robots, yet robust autonomy remains challenging on construction sites.
These environments are dynamic and often visually occluded, which complicates perception and navigation. In this context, valuable information from audio sources remains underutilized in most autonomy stacks. This work presents a control barrier function (CBF)-based safety filter that provides safety guarantees for obstacle avoidance while adapting safety margins during navigation using an audio-derived risk cue.
The proposed framework augments the CBF with a lightweight, real-time jackhammer detector based on signal envelope and periodicity. Its output serves as an exogenous risk that is directly enforced in the controller by modulating the barrier function. The approach is evaluated in simulation with two CBF formulations (circular and goal-aligned elliptical) with a unicycle robot navigating a cluttered construction environment.
Results show that the CBF safety filter eliminates safety violations across all trials while reaching the target in 40.2\% (circular) vs. 76.5\% (elliptical), as the elliptical formulation better avoids deadlock.
This integration of audio perception into a CBF-based controller demonstrates a pathway toward richer multimodal safety reasoning in autonomous robots for safety-critical and dynamic environments.
\end{abstract}

\section{Introduction}

With the growing field of construction robotics, the need for autonomy becomes more prominent. Industry demand is strong, but deployment remains constrained by safety requirements, low predictability, and the uniqueness of jobsites \cite{parascho2023construction}. 
Unlike controlled factory environments, construction takes place in an unstructured environment with conditions that evolve throughout the project, which complicates reliable on-site autonomy \cite{melenbrink2020site}. 
Robot autonomy enables tasks to be performed without human intervention and is a key step toward semi-autonomous or fully autonomous unmanned ground vehicles (UGVs) or unmanned aerial vehicles (UAVs) \cite{park2018automated}. Applications of autonomous robots include visualization and site preparation \cite{elmakis2022vision}.

While the field of robotics and control has made significant progress in other industrial applications, it remains to be widely adopted in construction. This presents an opportunity to integrate proven concepts, even though they must be adapted to the realities of construction-specific applications. Autonomous robotic vehicles rely on localization, mapping, sensing, and control, and planning \cite{lynch2017modern}. Obstacles can be bypassed through global replanning or local avoidance, this paper focuses on the latter.

Obstacle avoidance has long been studied, such as with virtual force fields and artificial potential fields \cite{borenstein1989real}. However, artificial potential fields can suffer from local minima and tuning-induced oscillations, while control barrier functions (CBFs) provide a minimally invasive safety filter that modifies the nominal controller only when required \cite{singletary2021comparative}.
CBFs have been thoroughly investigated in a variety of settings in the control research \cite{ames2019control}, yet they may be adjusted to the specifics of construction. 

Vision has been an important part of autonomous robotics in construction for quite some time \cite{feng2015vision}. There are many applications of visual sensor integration on construction sites such as for object detection, often including the use of ML-based methods.
Other approaches seek to use gas sensors \cite{cheung2018real} or audio activity recognition, mostly relying on machine learning methods such as SVMs \cite{rashid2020activity} or CNN-LSTMs \cite{mannem2024smart}. The objective is primarily regarded as a classification problem, aiming to assign correctly label sounds \cite{lee2020evidence}. Audio-informed navigation in construction robotics is comparatively underutilized. Most prior work uses audio for hazard detection and alerts rather than directly shaping robot motion. Especially on dynamic, cluttered, and noisy construction sites, additional contextual information can provide value.

This paper investigates two underserved areas in construction robotics: safety-aware control for site navigation beyond classic methods, and the integration of audio cues as additional context to modulate robot behavior during hazardous activity. This paper presents the following contributions:
\begin{itemize}
    \item The implementation and evaluation of a CBF safety filter for obstacle avoidance in a Building Information Modeling (BIM)-derived, construction-like environment.
    \item A goal-aligned elliptical barrier formulation that mitigates deadlock behaviors observed with circular CBFs.
    \item A lightweight audio detector using jackhammer sound that modulates the CBF boundary to integrate contextual risk.
\end{itemize}

\subsection{Control Barrier Functions}
Control barrier functions are used to describe obstacles and provide a control law on how to avoid them. A standard CBF is described as in \eqref{eq:standard_CBF}, where the safe condition is the inequality $h_1>0$, and $x_c$ and $y_c$ denote the center of an obstacle with radius $r$.

\begin{equation}
h_1 = (x - x_c)^2 + (y - y_c)^2 - r^2
\label{eq:standard_CBF}
\end{equation}

A common kinematic model for mobile robots is the unicycle model. Its nonlinear characteristics are depicted in the standard unicycle model in control-affine form $\dot s= f(s) + g(x)u$ with the state $\mathbf{s}=\big[x,y,\theta \big]^\top$ in \eqref{eq:standard_unicycle}, where $x$ is the forward position, $y$ is the lateral position, and $\theta$ is the yaw angle. Note that the state is influenced by two control inputs: linear velocity $v$ and angular velocity $\omega$.

\begin{equation}
\dot s = 
\begin{bmatrix}\dot x\\ \dot y\\ \dot \theta\end{bmatrix} =
\begin{bmatrix}\cos\theta\\ \sin\theta\\ 0\end{bmatrix} v + \begin{bmatrix}0\\0\\1\end{bmatrix} \omega
\label{eq:standard_unicycle}
\end{equation}

A common problem with the use of the CBF \eqref{eq:standard_CBF} for the standard unicycle model \eqref{eq:standard_unicycle} is that it does not provide control authority over both control inputs $v$ and $\omega$ to steer the robot. This is referred to as a mixed relative degree problem.
Due to this problem, it is beneficial to use an augmented unicycle model \eqref{eq:augmented_unicycle} as discussed in \cite{kim2025robust}, which increases the relative degree. The CBF can be modified via backstepping, which preserves the safety guarantees \cite{koga2023safe}. Backstepping enables to control both inputs. The forward velocity $v$ is then treated as a state rather than a control input in the augmented model \eqref{eq:augmented_unicycle} and is replaced by the forward acceleration $a$. With the new state $\mathbf{s}=\big[x,y, v,\theta \big]^\top$, the system now includes a drift, which is the part that is not controllable by the inputs.

\begin{equation}
\dot s =
\begin{bmatrix} \dot x \\ \dot y \\ \dot v \\ \dot \theta \end{bmatrix}
= \begin{bmatrix} v\cos\theta \\ v\sin\theta \\ 0 \\ 0 \end{bmatrix}
+ \begin{bmatrix} 0 \\ 0 \\ 1 \\ 0 \end{bmatrix} a
+ \begin{bmatrix} 0 \\ 0 \\ 0 \\ 1 \end{bmatrix} \omega
\label{eq:augmented_unicycle}
\end{equation}

The mixed relative degree problem can be addressed by defining the backstepping-based function $h_2$ in \eqref{eq:h2_definition}. $L_f h_1$ denotes the Lie derivative of $h_1$ with respect to $f$, $c_1$ is a tuning parameter and $h_1$ is the normalized CBF.

\begin{equation}
h_2 \;=\; c_1\,h_1 \;+\; L_f h_1 =  c_1\left(\frac{(x-x_c)^2+(y-y_c)^2}{r^2}-1\right)
+ v\left(
\frac{\partial h_1}{\partial x}\cos\theta
+\frac{\partial h_1}{\partial y}\sin\theta
\right) \\
\label{eq:h2_definition}
\end{equation}

With the backstepping-based CBF $h_2$, the safety condition can be formulated as in \eqref{eq:h2dot_inequality}. It requires an extended class-$\mathcal{K}$ function $\alpha(\cdot)$ and the Lie derivatives $L_f h_2$ and $L_g h_2$.
The Lie derivatives used in \eqref{eq:h2dot_inequality} are defined in \eqref{eq:Lg-h2_derivative} and \eqref{eq:Lf-h2_derivative}.

\begin{equation}
\dot h_2 \;=\; L_f h_2 \;+\; L_g h_2\,
\begin{bmatrix}a\\ \omega\end{bmatrix}
\;\ge\; -\,\alpha\!(h_2)
\label{eq:h2dot_inequality}
\end{equation}

\begin{equation}
L_g h_2
=
\begin{bmatrix}
\dfrac{\partial h_2}{\partial v} & \dfrac{\partial h_2}{\partial \theta}
\end{bmatrix}
=
\begin{bmatrix}
\displaystyle \frac{\partial h_1}{\partial x}\,\cos\theta
+ \frac{\partial h_1}{\partial y}\,\sin\theta
&
\displaystyle v\left(
-\frac{\partial h_1}{\partial x}\,\sin\theta
+ \frac{\partial h_1}{\partial y}\,\cos\theta
\right)
\end{bmatrix}
\label{eq:Lg-h2_derivative}
\end{equation}

\begin{equation}
L_f h_2
= v\cos\theta\,\frac{\partial h_2}{\partial x}
+ v\sin\theta\,\frac{\partial h_2}{\partial y}
\label{eq:Lf-h2_derivative}
\end{equation}

The CBF inequality \eqref{eq:h2dot_inequality} can be solved by a quadratic program (QP). The QP minimizes the deviation between the CBF control command $\mathbf{u}$ and a nominal control $\mathbf{u_{nom}}$. A nominal control describes the control used without obstacle avoidance that leads to a short but unsafe path. 
The QP solves the inequality \eqref{eq:qp_constraints} for $\mathbf{u} = [a \ \omega]^\top$ with bounds on the control inputs $u_i$. With the common choice of $\alpha(s)=c_2 s$ for some $c_2>0$, this becomes the linear inequality in the controls $(a,\omega)$ in \eqref{eq:qp_constraints}. When this inequality is satisfied, the robot is safe at all times if it was initialized in a safe set.

\begin{equation}
\begin{gathered}
L_g h_2
\begin{bmatrix} a \\ \omega \end{bmatrix}
\ge -L_f h_2 - c_2 h_2 \\[0.5em]
\min_{u\in\mathbb{R}^2}
(u-u_{\mathrm{nom}})^\top (u-u_{\mathrm{nom}}) \\
\text{s.t.}\;
u_{i,\min}\le u_i \le u_{i,\max}, \quad i=1,2
\end{gathered}
\label{eq:qp_constraints}
\end{equation}

The QP does not include a relaxation or slack variable. Consequently, the QP may become infeasible if the CBF inequality conflicts with actuator constraints or if the state-dependent unsafe set abruptly increases by external risk. Although the QP remained feasible in all tested scenarios, infeasibility remains a practical risk.

The beneficial property of safety comes along with the practical challenge of deadlock, where the robot faces an obstacle and is unable to navigate around it. 
This paper proposes elliptical CBFs for this purpose, as they can create more favorable conditions for smooth obstacle avoidance and reduce deadlocks, even though they expand the unsafe region.

The rationale in the derivation is the same as for circular CBFs. Unlike circular CBFs, they also allow the major and minor axes to be aligned, represented by the angle $\phi$. One way to do this is to use the direction toward the robot's next target $\phi=\operatorname{atan2}(y_{goal}-y_c,\;x_{goal}-x_c)$. By elongating and aligning the barrier along the target direction, the constraint preserves tangential feasible directions near the obstacle and reduces the tendency of circular CBFs to cause deadlock. This alternative formulation of the CBF $h_1$ is given in \eqref{eq:h1_ellipse}. The ellipse is centered at \((x_c, y_c)\) with semi-axes \(a>0\) and \(b>0\), rotated by an angle \(\phi\). The details of the backstepping-based $h_2$ and Lie derivatives are provided in Appendix A.

\begin{equation}
\begin{aligned}
h_1
&= \left(
\frac{\cos\phi\,\big(x - x_c\big) + \sin\phi\,\big(y - y_c\big)}{a}
\right)^{\!2}
+
\left(
\frac{-\sin\phi\,\big(x - x_c\big) + \cos\phi\,\big(y - y_c\big)}{b}
\right)^{\!2}
- 1
\end{aligned}
\label{eq:h1_ellipse}
\end{equation}

\section{Methodology}
 
 
This paper embeds contextual information directly into the controller by modulating the barrier. While vision is commonly used for localization and mapping, audio is less common. Here, it is used as a contextual cue to modulate local risk.
Risk-sensitive CBF formulations that incorporate uncertainty have been investigated, leading to a risk-sensitive barrier condition \cite{singletary2022safe}.
This paper treats the output of an audio hazard detector as an exogenous risk cue that parameterizes the barrier constraint.
As a proxy, this paper considers the audio signal of human workers using jackhammers, which can pose risk to both humans and robots in a collaborative environment. Note that other audio signals are also possible and are not limited to jackhammers.

The lightweight jackhammer sound detector does not rely on machine learning models or the integration of LLMs, which enables low latency and transparency.
The jackhammer detector uses two cues: signal-to-noise ratio (SNR) and periodicity (autocorrelation). SNR captures on/off amplitude changes by computing the RMS of short signal  bursts of the signal $x$ over $N$ samples, then estimating an adaptive noise floor as the rolling-window median over $M$ samples. The SNR is computed in decibels in \eqref{eq:SNR}.
Periodicity is computed via autocorrelation. Not every volume change is caused by jackhammer sounds. The autocorrelation computes the prevailing frequency of the signal. Repeated pulses in the relevant frequency band indicate active jackhammer sounds. The value $f_{peak}$ is taken as the maximum magnitude at a certain frequency in \eqref{eq:autocorr}.

\begin{equation}
\mathrm{SNR}_{\mathrm{dB}} = 10\log_{10}\!\left(\frac{E_{current}}{E_{floor}}\right),
\qquad
E_{\text{current}}=\sqrt{\frac{1}{N}\sum_{i=1}^{N} x_i^{2}},
\qquad
E_{\text{floor}}=\operatorname*{median}_{k\in\{1,\ldots,M\}}\!\left(E_{t-k}\right)
\label{eq:SNR}
\end{equation}

\begin{equation}
f_{peak} = max(\rho(\tau)),
\quad
\rho(\tau)=
\frac{\sum_{t}\big(e_t-\bar e\big)\big(e_{t+\tau}-\bar e\big)}
{\sum_{t}\big(e_t-\bar e\big)^2},
\quad
\bar e=\frac{1}{K}\sum_{t=1}^{K} e_t,
\quad
e_t=\sqrt{\frac{1}{N}\sum_{i=1}^{N} x_i^2}
\label{eq:autocorr}
\end{equation}

The two components are fused into a decision logic. The classifier switches the status $z(t)$ to ON/OFF under the following conditions: $\text{z(t)=ON if} \ \mathrm{SNR}_{\mathrm{dB}}\ge \mu_{\mathrm{on}}\ \wedge\ f_{peak}\ge \gamma_{\mathrm{on}}$. 
Also, $\text{z(t)=OFF if} \  \mathrm{SNR}_{\mathrm{dB}}< \mu_{\mathrm{off}}\ \wedge\ f_{peak}< \gamma_{\mathrm{off}}$.

Risk is mapped to geometry by modifying the CBF boundary online. Each obstacle stores a baseline risk level $d_{\text{base}}\ge 0$.
Given an audio-derived risk status $z(t)$, the radius $r$ is inflated by $r_{new}(t)=r+d_{\text{base}}\,z(t)$. For elliptical CBFs, both axes are inflated accordingly. The detected risk automatically inflates all obstacles, since assignment to individual sources is not within the scope here.

This selection is validated with four sound sources that capture the sound of a jackhammer and background noise. Thresholds are found heuristically.
Note that introducing changing radii leads to a state-dependent CBF. While this can place the robot within the unsafe set, it reduces the meaningfulness of the formal CBF guarantees, since the robot can effectively be initialized within the unsafe area. Thus, audio modifies the conservativeness of the safety filter, and this paper evaluates collision reduction and clearance empirically rather than formal guarantees.

\section{Simulation study}
The simulation of the control approach takes place in an environment with walls and two temporary structure obstacles modeled in BIM within a $5.5\,\mathrm{m}$ by $10\,\mathrm{m}$ area. Two scenarios are investigated: the first involves an agent driving in a straight line around two obstacles. In the second scenario, the robot must traverse six waypoints in a construction-like environment. The obstacles are inflated corresponding to a simultaneous jackhammer detection.
Each scenario is evaluated with three control modes: nominal control, circular CBF, and elliptical CBF.
The nominal control $\mathbf{u_{nom}} = [a_{nom} \ \omega_{nom}]^\top$ is used for comparison without obstacle avoidance in \eqref{eq:nominal_control} with $v_{des}$ as desired velocity and $\theta_{des}$ as desired angle to the next waypoint.
Note that this second-order control of the state can pose challenges for robots using only first-order commands, which is why the acceleration control signal is integrated.
The elliptical CBFs are defined with an aspect ratio of 1.8.
Table \ref{table:simulation_parameters} shows the parameters of the control and simulation.

\begin{equation}    
a_{nom} = -k_1\,(v - v_{des}),
\qquad
\omega_{nom} = -k_2(\theta - \theta_{des})
\label{eq:nominal_control}
\end{equation}

\begin{table}[!htbp]
\caption{Parameters used for the CBF–QP safety filter and the simulation study}
\label{table:simulation_parameters}
\centering
\small
\renewcommand{\arraystretch}{1.25}
\begin{tabular}{l l | l l}
\hline\hline
\multicolumn{1}{c}{Parameter} & \multicolumn{1}{c}{Value} &
\multicolumn{1}{c}{Parameter} & \multicolumn{1}{c}{Value} \\
\hline
CBF parameter $c_1$ & 3.0 & Nominal control parameter $k_1$ & 1.0 \\
CBF parameter $c_2$ & 1.0 & Nominal control parameter $k_2$ & 1.0 \\
Simulation time step $dt$ & 0.1 & Max. abs. linear acceleration $a_{max} \ [m/s^2]$ & 2.5 \\
Desired velocity $v_{des}$ \ [m/s] & 0.5 & Max. abs. angular velocity $\omega_{max} \ [rad/s]$ & 1.0 \\
\hline\hline
\end{tabular}
\normalsize
\end{table}

The values for the jackhammer detector are $\mu_{\mathrm{on}} = 2\,\mathrm{dB}$ and $\mu_{\mathrm{off}} = -1\,\mathrm{dB}$, as well as $\gamma_{\mathrm{on}} = 0.22$ and $\gamma_{\mathrm{off}} = 0.12$, empirically determined over four different audio signals. The on-delay is $180\,\mathrm{ms}$ and the off-delay is $400\,\mathrm{ms}$.

The following metrics are reported: path length, minimal signed distance to obstacle, safety violation time, completion time, and success rate. A negative minimum signed distance implies a safety violation. Safety violation time is the cumulative duration during in which the CBF inequality is violated, and the success rate denotes the reach of the target. The location of obstacles is randomly perturbed by shifting their position with offsets $d_x, d_y \in (-2r, 2r)$.
Power analysis showed a necessary sample size of $N=14.16$ for a precision of $0.025\,\mathrm{m}$ for distance to obstacle. The experiment is repeated $N=15$ times, and the mean and standard deviation of all successful results are reported.

\section{Results}

\subsection{Scenario 1: Elliptical, goal-aligned CBF}
\begin{figure}[!htbp]
\centering
\begin{minipage}[t]{0.33\linewidth}
  \centering
  \includegraphics[width=\linewidth]{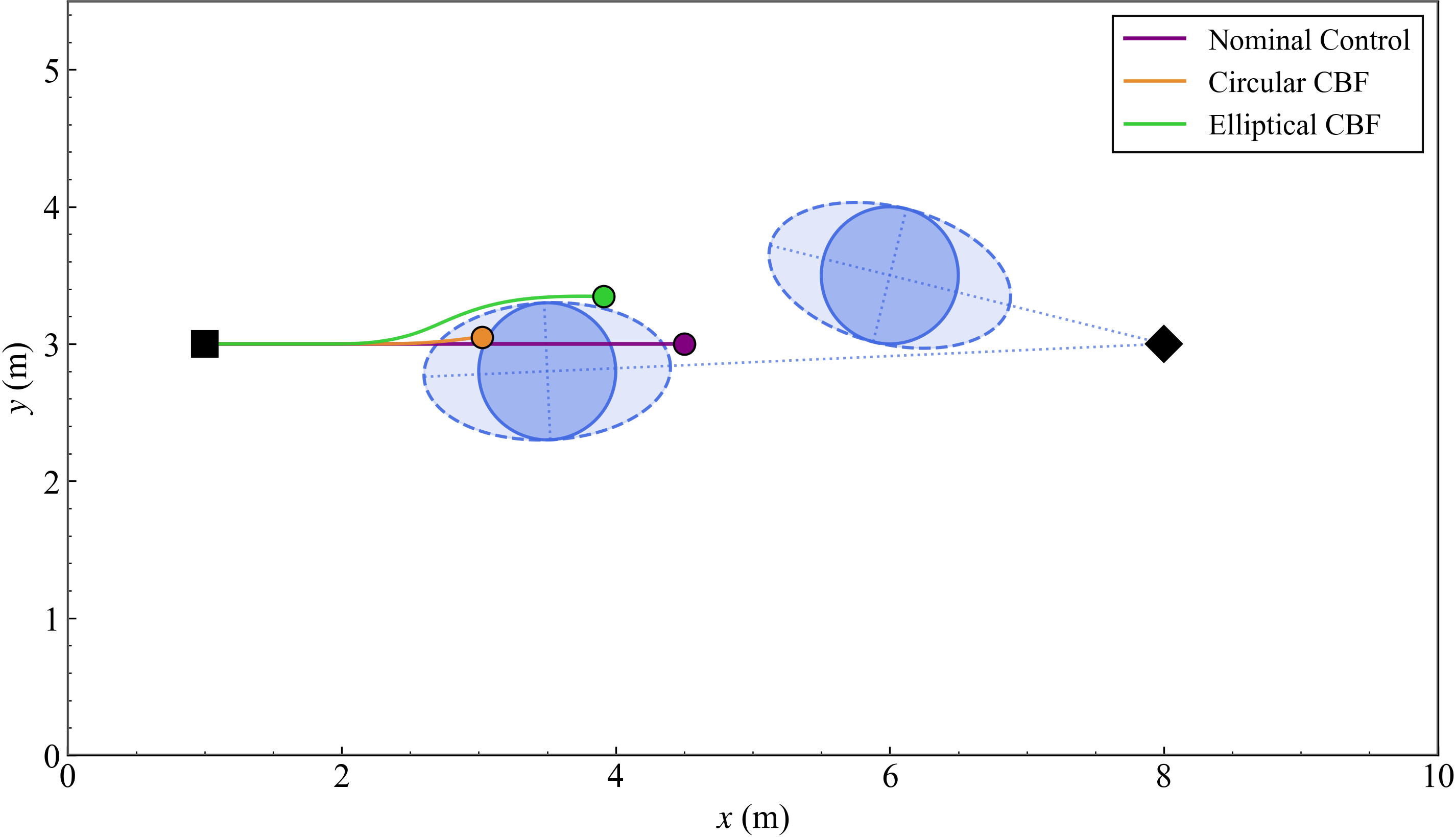}
  \caption*{(a)}
\end{minipage}\hfill
\begin{minipage}[t]{0.33\linewidth}
  \centering
  \includegraphics[width=\linewidth]{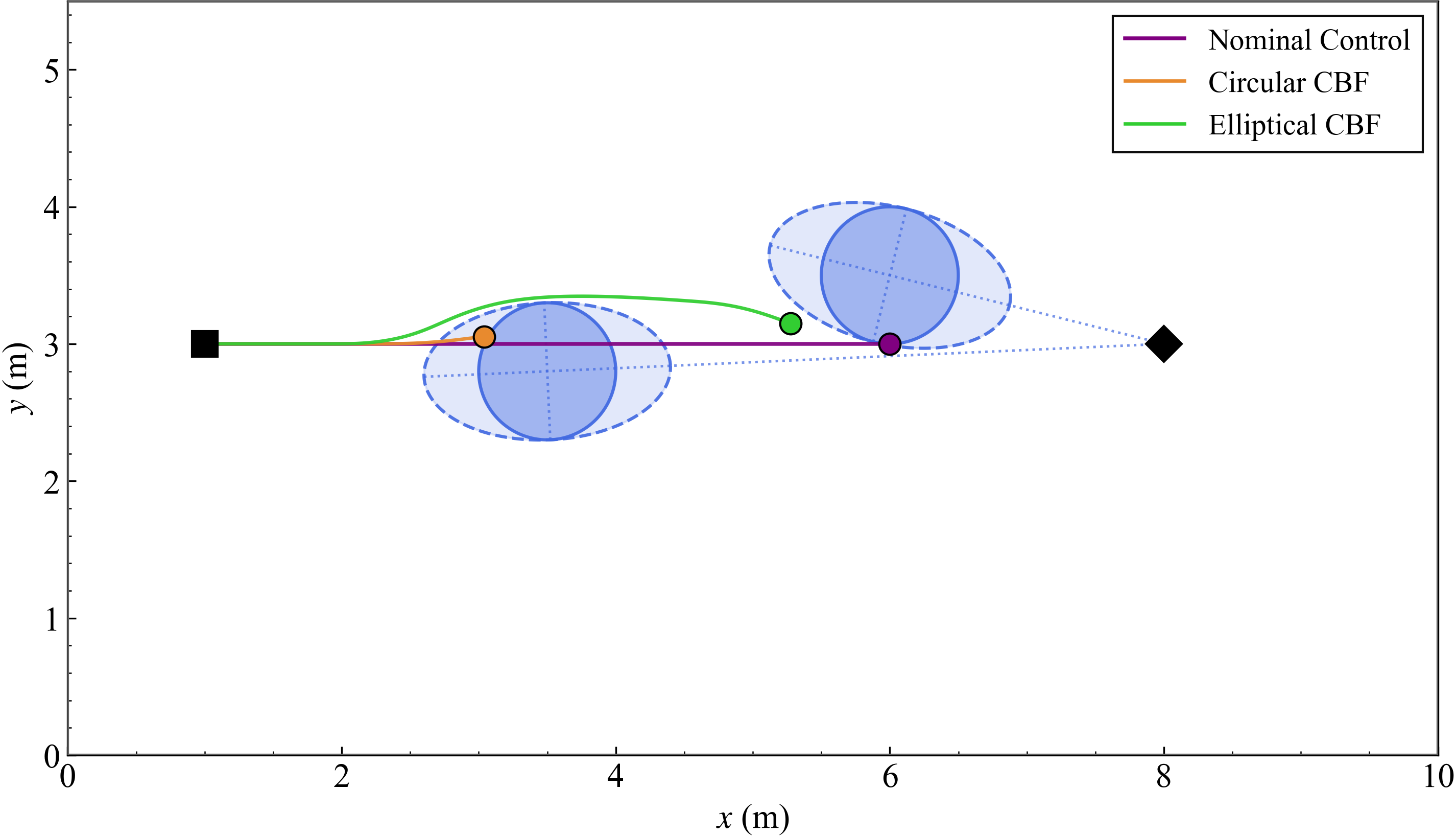}
  \caption*{(b)}
\end{minipage}
\begin{minipage}[t]{0.33\linewidth}
  \centering
  \includegraphics[width=\linewidth]{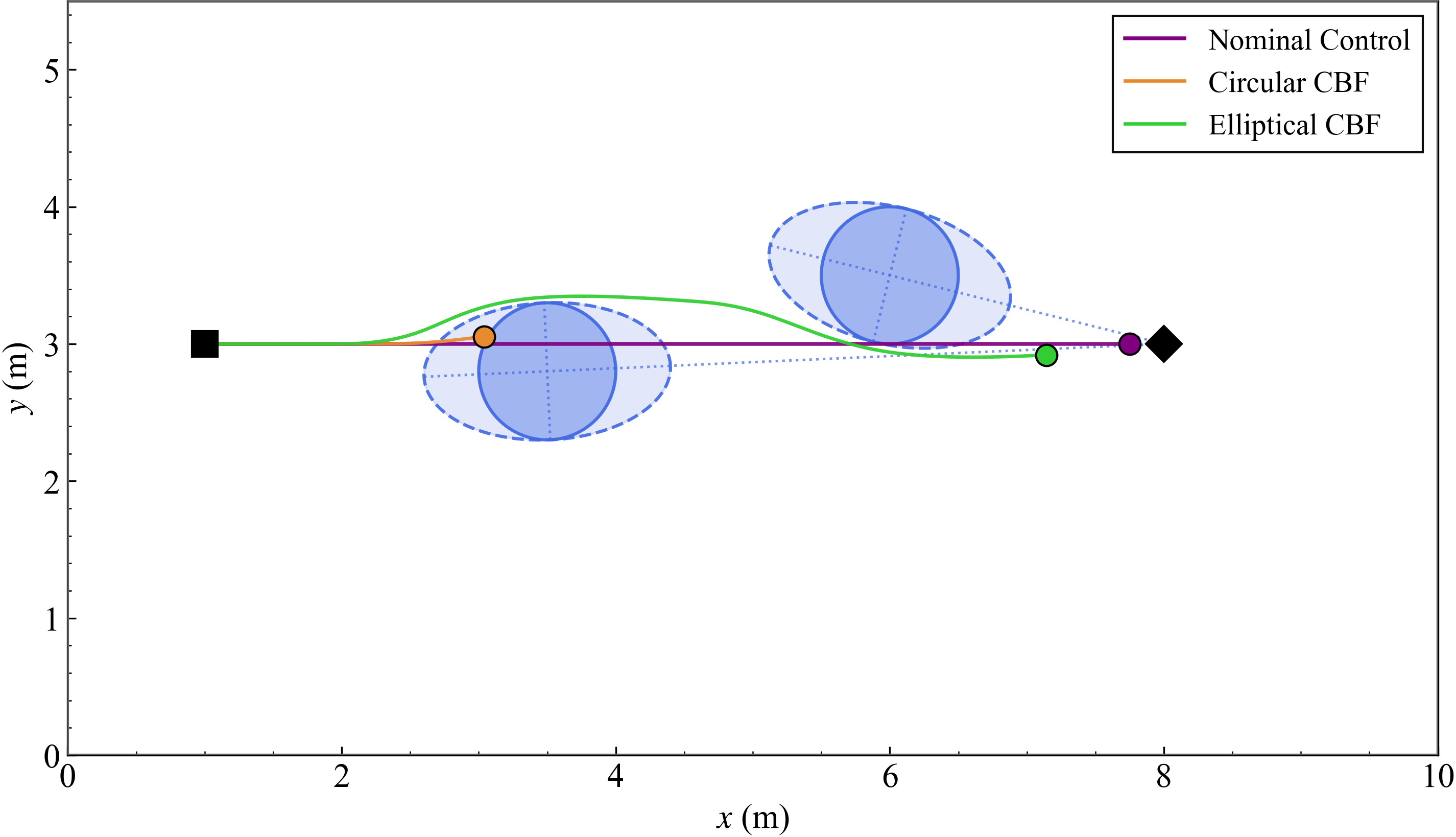}
  \caption*{(c)}
\end{minipage}
\caption{Scenario~1: Robot trajectories for the three control strategies at three time steps: (a) $t_1 = 8.0\,\mathrm{s}$ first obstacle encounter, (b) $t_2 = 11.0\,\mathrm{s}$ second obstacle encounter, and (c) $t_3 = 15.0\,\mathrm{s}$ approach to the target. While the nominal control does not avoid obstacles, the circular CBF leads to deadlock and the goal-aligned elliptical CBF navigates safely to the target.
}
\label{fig:trajectory_scene1_all}
\end{figure}

This first set of experiments shows the robot navigating both circular and elliptical obstacles. The results demonstrate that using elliptical CBFs is more beneficial than using circular CBFs because the system is less prone to deadlock.  Fig.~\ref{fig:trajectory_scene1_all} shows an example trial and Table~\ref{table:cbf_metrics_scene1} reports the metrics over $N=15$.

\begin{table}[!htbp]
\caption{Scenario~1: Performance metrics over $N=15$ trials (mean $\pm$ std), computed over successful trials.}
\label{table:cbf_metrics_scene1}
\centering
\small
\renewcommand{\arraystretch}{1.25}
\setlength{\tabcolsep}{3pt}
\begin{tabular}{l c c c c c}
\hline\hline
\multicolumn{1}{c}{\makecell{Method\\}} &
\multicolumn{1}{c}{\makecell{Path\\Length [m]}} &
\multicolumn{1}{c}{\makecell{Minimum Signed\\Distance {[m]}}} &
\multicolumn{1}{c}{\makecell{Safety Violation\\Time [s]}} &
\multicolumn{1}{c}{\makecell{Completion \\Time [s]}} &
\multicolumn{1}{c}{\makecell{Success\\Rate}} \\
\hline
Nominal Control          & $6.750 \pm 0.000$ & $-0.240 \pm 0.166$ & $1.75 \pm 1.07$ & $14.40 \pm 0.00$ & $15/15 \ (100\%)$  \\
Circular CBF (static)    & $6.811 \pm 0.031$ & $0.080 \pm 0.044$  & $0.00 \pm 0.00$ & $16.20 \pm 1.06$ & $7/15 \ (47\%)$  \\
Elliptical CBF (static)  & $6.828 \pm 0.054$ & $0.054 \pm 0.017$  & $0.00 \pm 0.00$ & $15.30 \pm 0.84$ & $11/15 \ (73\%)$  \\
\hline\hline
\end{tabular}
\normalsize
\end{table}

\subsection{Scenario 2: Audio-parametrized CBF}
\begin{figure}[!htbp]
\centering
\begin{minipage}[t]{0.33\linewidth}
  \centering
  \includegraphics[width=\linewidth]{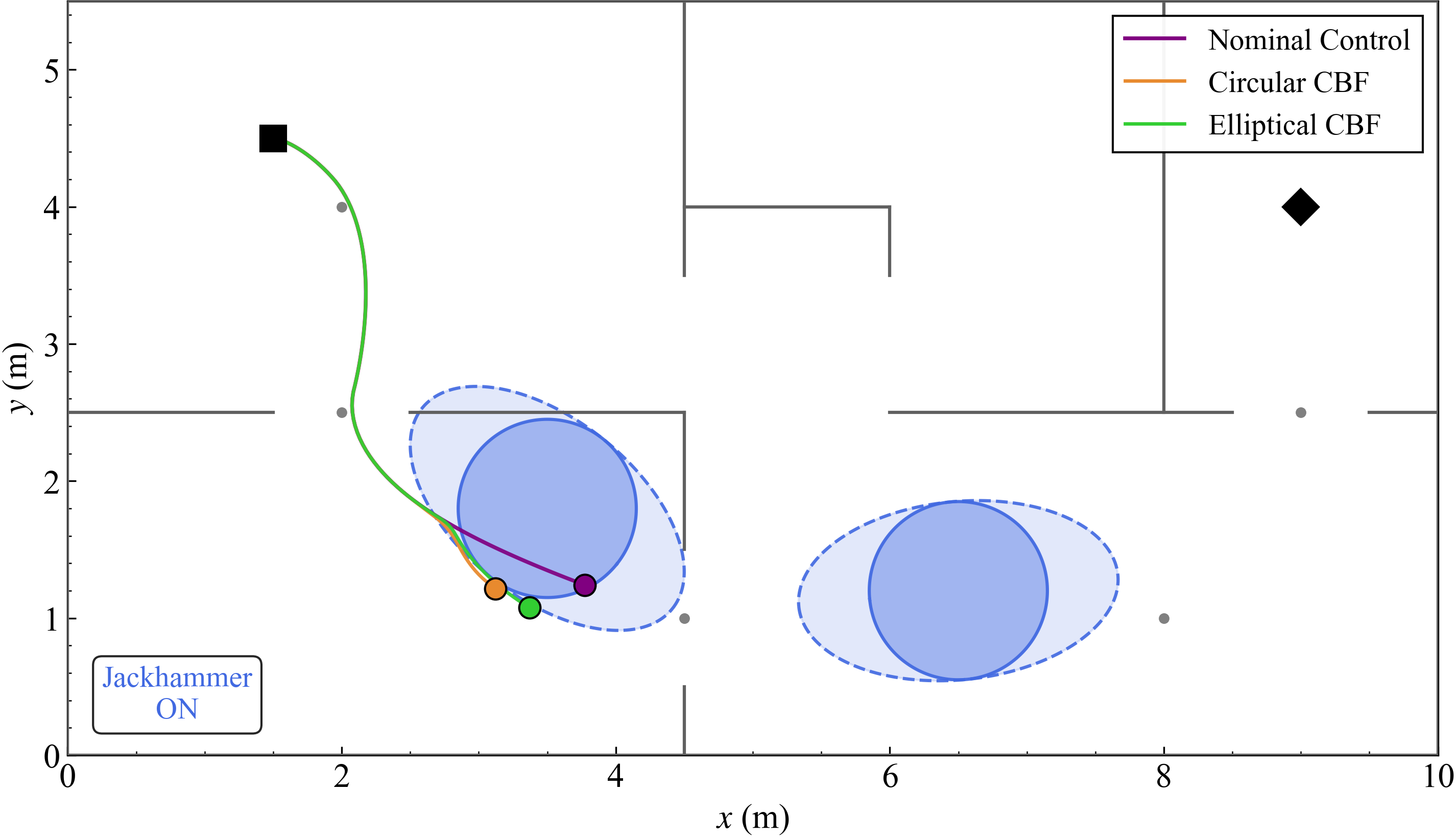}
  \caption*{(a)}
\end{minipage}\hfill
\begin{minipage}[t]{0.33\linewidth}
  \centering
  \includegraphics[width=\linewidth]{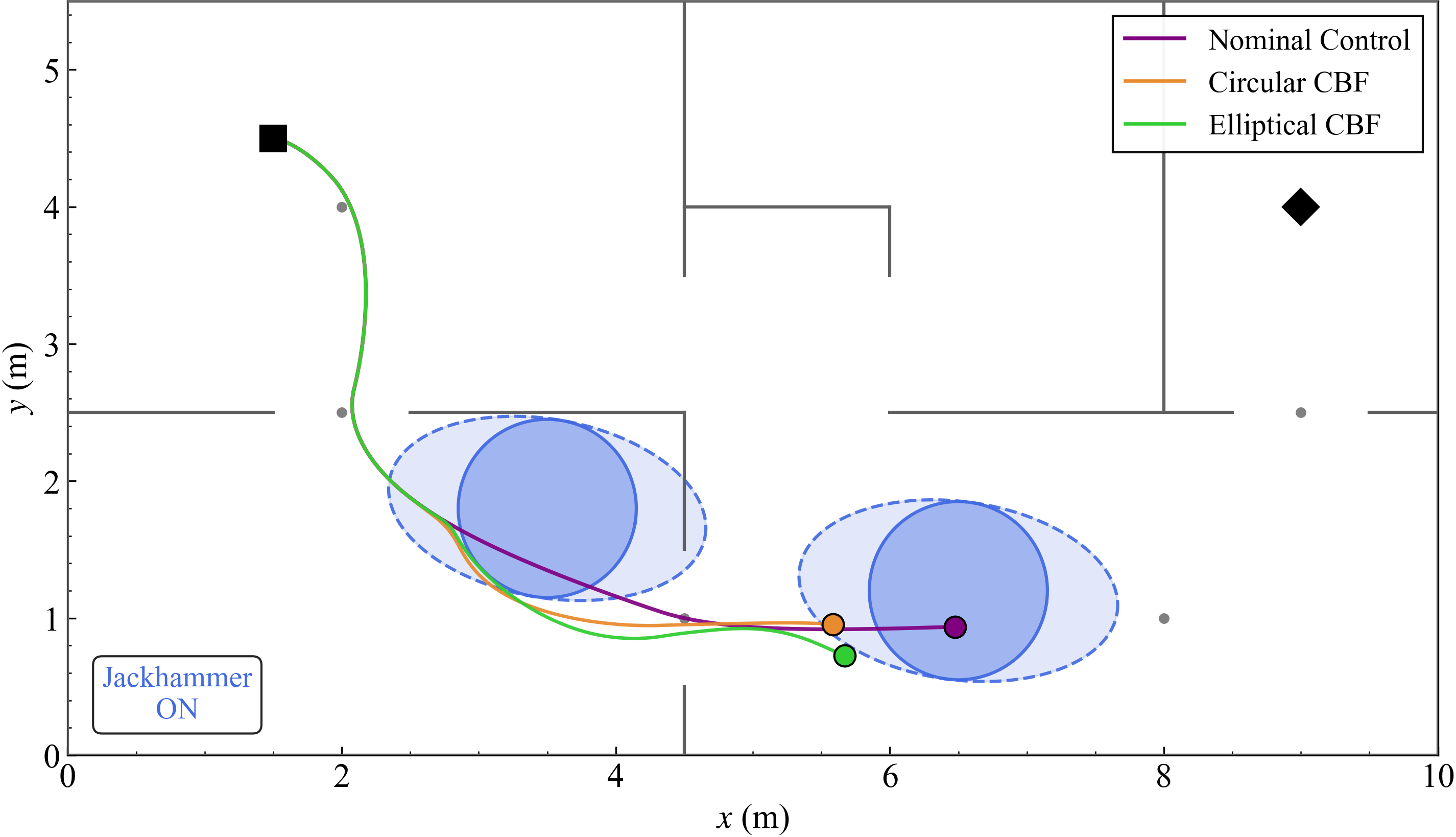}
  \caption*{(b)}
\end{minipage}
\begin{minipage}[t]{0.33\linewidth}
  \centering
  \includegraphics[width=\linewidth]{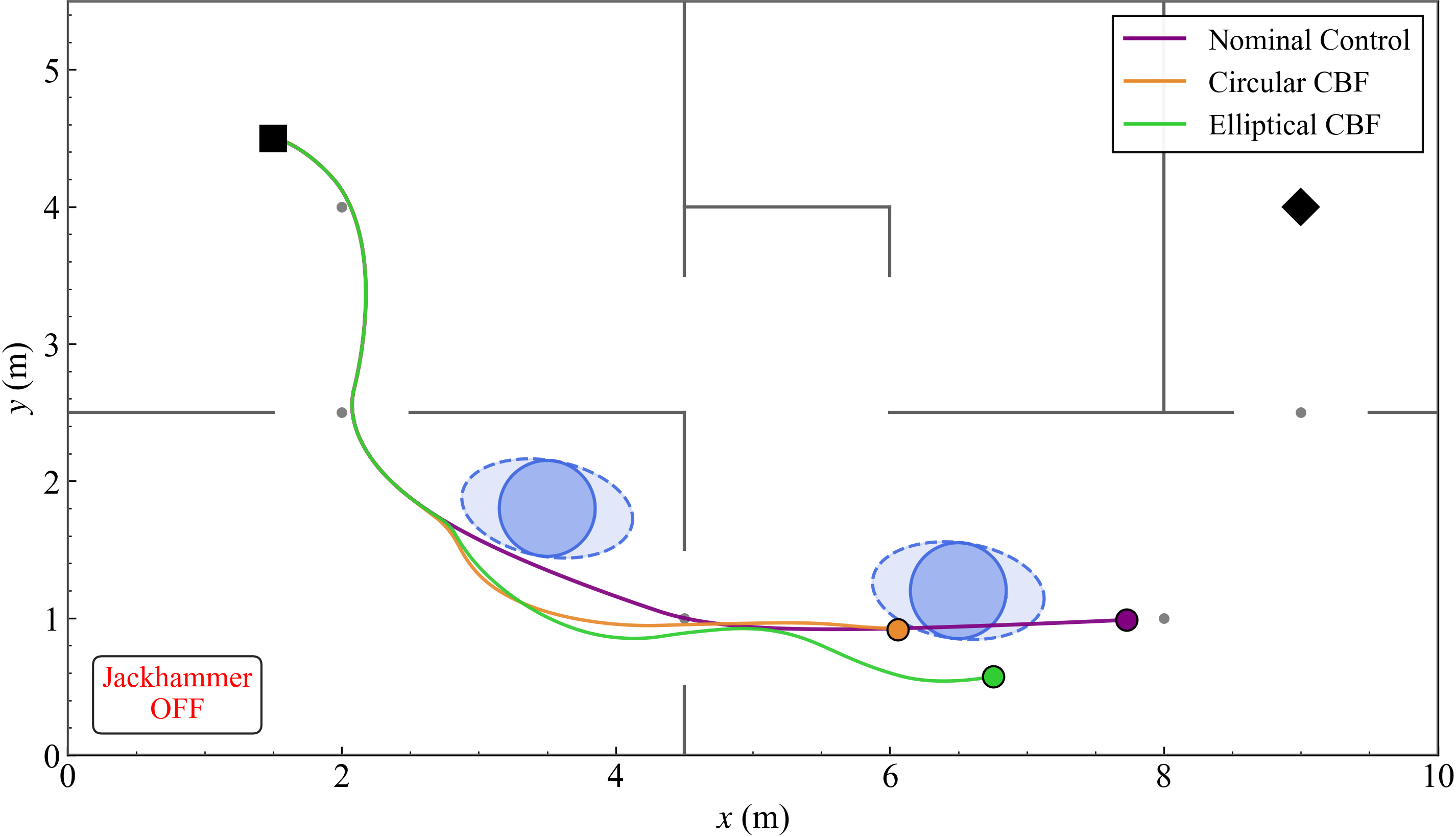}
  \caption*{(c)}
\end{minipage}
\caption{Scenario~2: Robot trajectories for the case with audio-aware CBF modulation. (a) $t_1 = 10.0\,\mathrm{s}$ first obstacle encounter with inflated obstacle under active jackhammer detection, (b) $t_2 = 15.5\,\mathrm{s}$ second obstacle encounter, and (c) $t_3 = 18.0\,\mathrm{s}$ decrease in detected risk leading to a deflated obstacle.}
\label{fig:trajectory_scene2_all}
\end{figure}

The second scenario shows the robot operating with a state-dependent CBF, the radius of which is modified in the described audio-parametrized way. An example trajectory with three time steps is shown in Fig.~\ref{fig:trajectory_scene2_all}. Quantitative results are reported in Table~\ref{table:cbf_metrics_scene2}.

\begin{table}[!htbp]
\caption{Scenario~2: Performance metrics over $N=15$ trials (mean $\pm$ std), computed over successful trials.
}
\label{table:cbf_metrics_scene2}
\centering
\small
\renewcommand{\arraystretch}{1.25}
\setlength{\tabcolsep}{3pt}
\begin{tabular}{l c c c c c}
\hline\hline
\multicolumn{1}{c}{\makecell{Method\\}} &
\multicolumn{1}{c}{\makecell{Path\\Length [m]}} &
\multicolumn{1}{c}{\makecell{Minimum Signed\\Distance {[m]}}} &
\multicolumn{1}{c}{\makecell{Safety Violation\\Time [s]}} &
\multicolumn{1}{c}{\makecell{Completion \\Time [s]}} &
\multicolumn{1}{c}{\makecell{Success\\Rate}} \\
\hline
Nominal Control          & $11.900 \pm 0.000$ & $-0.170 \pm 0.169$ & $1.57 \pm 1.09$ & $24.70 \pm 0.00$ & $15/15 \ (100\%)$ \\
Circular CBF (audio)    & $11.947 \pm 0.032$ & $0.130 \pm 0.067$  & $0.00 \pm 0.00$ & $25.76 \pm 0.95$ & $5/15 \ (33.3\%) $ \\
Elliptical CBF (audio)  & $12.033 \pm 0.074$ & $0.075 \pm 0.057$  & $0.00 \pm 0.00$ & $25.57 \pm 0.55$ & $12/15 \ (80\%)$ \\
\hline\hline
\end{tabular}
\normalsize
\end{table}

\begin{figure}[!htbp]
\centering
\hspace*{\fill}
\begin{minipage}[t]{0.4\linewidth}
  \centering
  \includegraphics[width=\linewidth]{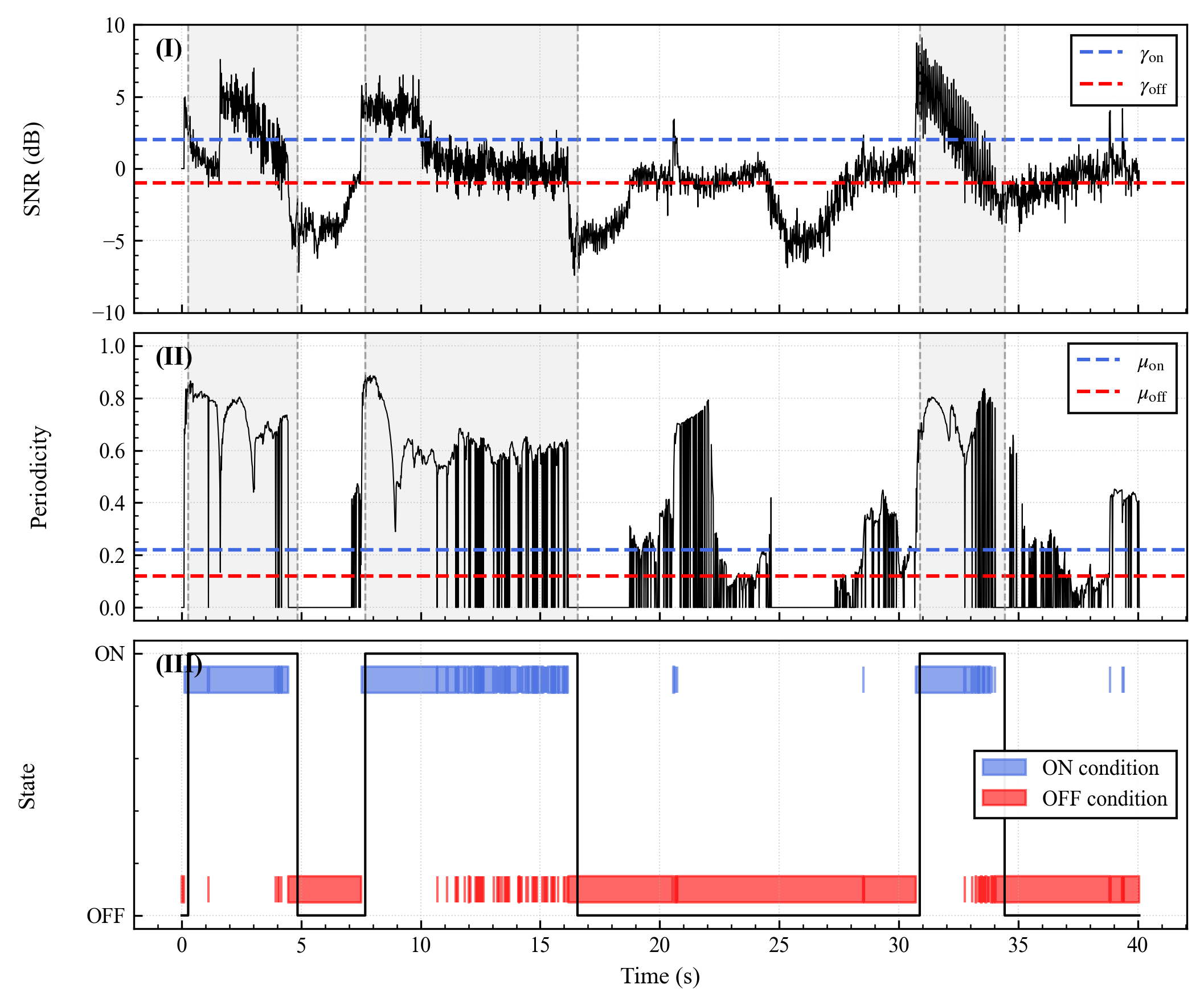}
  \caption*{(a)}
\end{minipage}\hfill
\begin{minipage}[t]{0.4\linewidth}
  \centering
  \includegraphics[width=\linewidth]{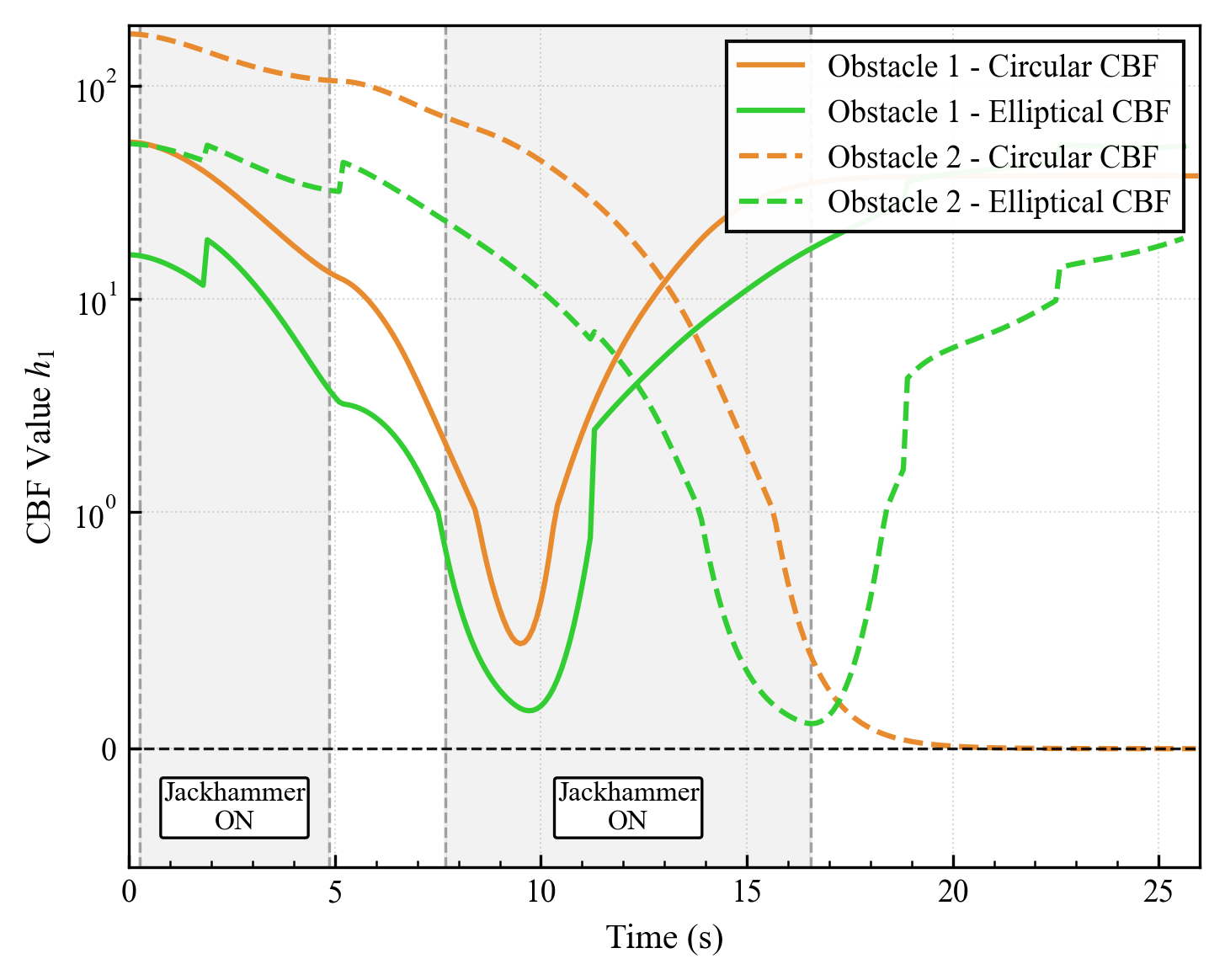}
  \caption*{(b)}
\end{minipage}
\hspace*{\fill}
\caption{(a) Jackhammer detector mechanism: (I) Signal-to-noise ratio with ON/OFF thresholds, (II) Periodicity from autocorrelation with thresholds, (III) State of the detector, ambiguities are resolved by the decision logic. 
(b) CBF values $h_1>0$ indicating the safety of each method for both obstacles. The gray background indicates the time periods when the jackhammer detector was turned on.}
\label{fig:jackhammer_detector_cbf_values}
\end{figure}

The effect of the jackhammer detector is shown in Fig.~\ref{fig:jackhammer_detector_cbf_values} (a). Fig  \ref{fig:jackhammer_detector_cbf_values} (b) shows the change in $h_1$ over time under the influence of audio cues.

\section{Discussion}

Fig.~\ref{fig:trajectory_scene1_all} shows an exemplary trial from Scenario~1, demonstrating a common case where the circular CBF leads to deadlock. In contrast, the elliptical CBF allows the robot to reach the target.
Table~\ref{table:cbf_metrics_scene1} summarizes the performance of the proposed approach. While nominal control generates the fastest and shortest path and reaches the goal in 100\% of trials, it fails by colliding with obstacles multiple times, resulting in unsafe behavior for $1.75\,\mathrm{s}$ and a negative minimum signed distance to obstacles. Circular CBFs succeed in 47\% of trials, with the drawback of frequent deadlocks. Elliptical CBFs succeed in 73\% of trials, with only a marginally increased path length compared to circular CBFs. Neither circular nor elliptical CBFs show safety violations.

Fig.~\ref{fig:trajectory_scene2_all} shows the performance of Scenario~2 with the proposed approach under audio-informed modulation of the CBF boundary. The circular CBF leads to a deadlock at the second obstacle. Results in Table~\ref{table:cbf_metrics_scene2} show an 80\% success rate for the elliptical CBF compared to 33.3\% for the circular CBF. The path length is slightly increased, while completion time is lower. Neither circular nor elliptical CBFs show safety violations.

Note also that the audio signal introduces a certain degree of ambiguity. Fig.~\ref{fig:jackhammer_detector_cbf_values}(a) the raw detector produces six on/off transitions that shift the CBF boundary, subfigure (III) shows how the decision logic suppresses this ambiguity and yields a stable risk state.
Fig.~\ref{fig:jackhammer_detector_cbf_values}(b) shows how $h_1$ values change for each obstacle, preserving safety at all times.

This approach shows potential for obstacle avoidance with CBFs on mobile robots. It demonstrates resilience to noise in a construction-like simulation environment. The proposed elliptical CBFs perform better than circular CBFs, as they provide greater resilience against deadlock. The context-aware, audio-based jackhammer detector integrates reliably and modulates the CBF condition directly within the controller. The mechanism with a self-adjusting major axis that targets the next waypoint integrates well with the objective guiding the robot more smoothly.

\section{Conclusion}

On-site autonomous construction remains limited because environments are unstructured, change over time, and contain the persistent challenge of robust on-site sensing. Many site hazards are context-dependent and may not be identifiable from visual signals alone, which motivates using additional context signals alongside geometric perception.
This work can be interpreted as evidence that construction audio signals, typically used for hazard detection, can be incorporated at the control layer as a contextual risk cue by directly injecting it into the safety constraint.
 
Some limitations can be identified in this paper. First, the lack of real-world implementation: the approach has only been demonstrated conceptually in simulation so far, and this will be addressed in future work. In particular, it remains to be investigated how well the formal guarantees hold under real-world disturbances. In addition, QPs are known to perform poorly when posed with infeasible constraints. Also, the conversion of LiDAR point clouds to geometric obstacles must be considered.

Second, the effect of jackhammer detection is evaluated using pre-recorded audio files. This can be further investigated to assess effects such as acoustic reflections and latency, which are not suitable for study in simulation. In addition, other classes of audio-based activation can be considered.

Third, there is no replanning implemented. This work focuses on the specific aspect of audio-informed CBFs, ignoring global replanning mechanisms that introduce new waypoints. The proposed scenarios assume a coarse representation, such as when only the centers of rooms and doors are known, rather than a finely discretized grid.

Fourth, efforts can be made to retain CBF guarantees under time-variant or state-dependent conditions when considering the formal guarantees. This would come with additional constraints, but it might be worth evaluating. Promising candidates include adaptive CBFs, although they are more demanding to implement.
However, these limitations can be addressed and they offer opportunities for both theoretical advances in control and practical deployment in construction robotics.

\bibliography{ascexmpl-new}

\section{Appendix A}

The CBF for the elliptical case with a rotated axis is described by $h_1$ in \eqref{eq:h1_ellipse} and $h_2$ in \eqref{eq:h2_elliptical}. The Lie derivative $L_f h_2$ is given in \eqref{eq:Lf-h2_elliptical}, and $L_g h_2$ is the same as in \eqref{eq:Lg-h2_derivative}.

\begin{equation}
\label{eq:h2_elliptical}
\begin{aligned}
h_2
&= c_1\Bigg\{
\left(
\frac{\cos\phi\,\big(x - x_c\big) + \sin\phi\,\big(y - y_c\big)}{a}
\right)^{\!2}
+
\left(
\frac{-\sin\phi\,\big(x - x_c\big) + \cos\phi\,\big(y - y_c\big)}{b}
\right)^{\!2}
- 1
\Bigg\} \\[6pt]
&\quad
+ 2v\Bigg\{
\Bigg[
\left(\frac{\cos^2\phi}{a^2}+\frac{\sin^2\phi}{b^2}\right)\,\big(x-x_c\big)
+
\left(\frac{\cos\phi\sin\phi}{a^2}-\frac{\cos\phi\sin\phi}{b^2}\right)\,\big(y-y_c\big)
\Bigg]\cos\theta \\[2pt]
&\qquad+
\Bigg[
\left(\frac{\cos\phi\sin\phi}{a^2}-\frac{\cos\phi\sin\phi}{b^2}\right)\,\big(x-x_c\big)
+
\left(\frac{\sin^2\phi}{a^2}+\frac{\cos^2\phi}{b^2}\right)\,\big(y-y_c\big)
\Bigg]\sin\theta
\Bigg\}
\end{aligned}
\end{equation}

\begin{equation}
\label{eq:Lf-h2_elliptical}
\begin{aligned}
L_f h_2
&= v\,c_1\!\left(
\frac{\partial h_1}{\partial x}\,\cos\theta
+ \frac{\partial h_1}{\partial y}\,\sin\theta
\right) 
+ v^2\!\left(
\frac{\partial^2 h_1}{\partial x^2}\cos^2\theta
+ 2\,\frac{\partial^2 h_1}{\partial x\,\partial y}\sin\theta\cos\theta
+ \frac{\partial^2 h_1}{\partial y^2}\sin^2\theta
\right)
\end{aligned}
\end{equation}

\end{document}